\documentclass[10pt,twocolumn,letterpaper]{article}

\usepackage{3dv}
\usepackage{times}
\usepackage{epsfig}
\usepackage{graphicx}
\usepackage{amsmath}
\usepackage{amssymb}

\usepackage{color} 
\usepackage{tabularx}
\usepackage{makecell}
\usepackage{multicol}
\usepackage{mwe,lipsum}
\usepackage{array,multirow}
\usepackage{float}
\usepackage{eucal}
\usepackage{soul}

\usepackage[pagebackref=true,breaklinks=true,colorlinks,bookmarks=false]{hyperref}

\threedvfinalcopy 

\def\threedvPaperID{11} 

\newcommand{\gr}[1]{{\textcolor{green}{\textbf{#1}}}}

\ifthreedvfinal\fi
\begin{document}

\title{Robust RGB-D Fusion for Saliency Detection }

\author{Zongwei Wu$^{1,2}$
\quad
Shriarulmozhivarman Gobichettipalayam$^{1}$
\quad
Brahim Tamadazte$^{3}$
\and
Guillaume Allibert$^{4}$
\quad
Danda Pani Paudel$^{2}$
\quad
C\'edric Demonceaux$^{1}$
\\
\normalsize ${}^{1}$ ImViA, Universit\'e Bourgogne Franche-Comt\'e, France \quad
\normalsize ${}^{2}$ Computer Vision Laboratory, ETH Zurich, Switzerland\\
\normalsize ${}^{3}$ Sorbonne Universit\'e, CNRS, ISIR, France \quad
\normalsize ${}^{4}$ Universit\'e  C\^ote d’Azur, CNRS, I3S, France
}

\maketitle
\begin{abstract}
Efficiently exploiting multi-modal inputs for accurate RGB-D saliency detection is a topic of high interest. Most existing works leverage cross-modal interactions to fuse the two streams of RGB-D for intermediate features' enhancement. In this process, a practical aspect of the low quality of the available depths has not been fully considered yet. In this work, we aim for RGB-D saliency detection that is robust to the low-quality depths which primarily appear in two forms: inaccuracy due to noise and the misalignment to RGB.  To this end, we propose a robust RGB-D fusion method that benefits from  (1) layer-wise, and (2) trident spatial, attention mechanisms. On the one hand, layer-wise attention (LWA) learns the trade-off between early and late fusion of RGB and depth features, depending upon the depth accuracy. On the other hand, trident spatial attention (TSA) aggregates the features from a wider spatial context to address the depth misalignment problem. 
The proposed LWA and TSA mechanisms allow us to efficiently exploit the multi-modal inputs for saliency detection while being robust against low-quality depths. Our experiments on five benchmark datasets demonstrate that the proposed fusion method performs consistently better than the state-of-the-art fusion alternatives. The source code is publicly available at:  \url{https://github.com/Zongwei97/RFnet}. 

\end{abstract}

\section{Introduction}
\begin{figure}[t]
\centering
\includegraphics[width=0.98\linewidth,keepaspectratio]{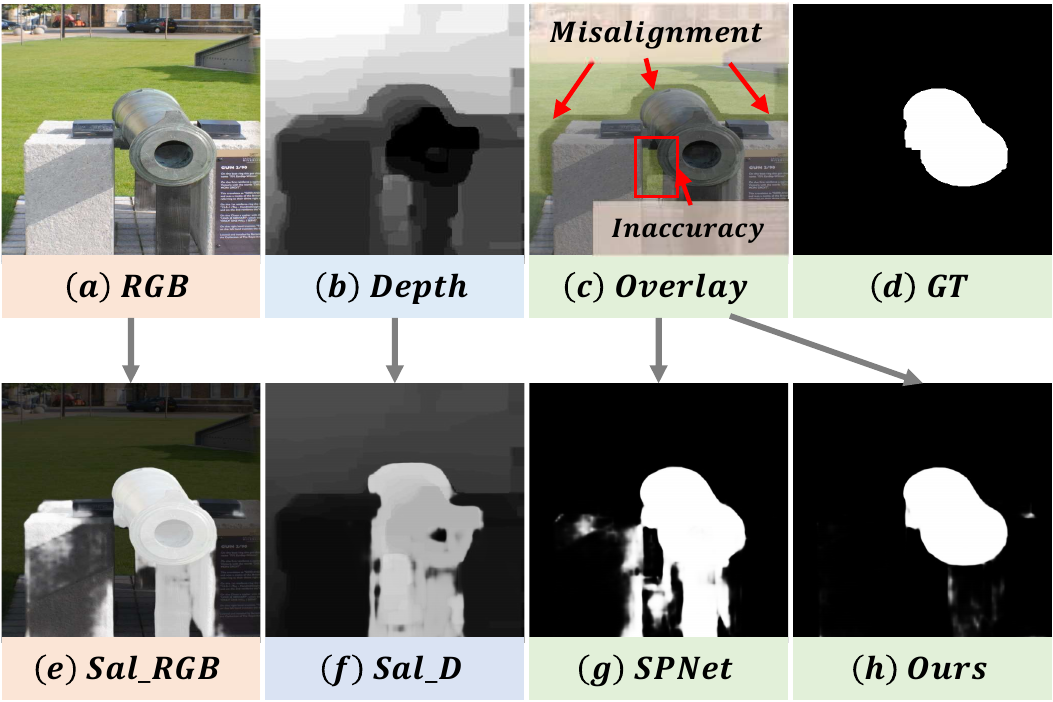}
\caption{\textbf{Motivation}. (a) and (b) are the paired RGB-D inputs. (e) and (f) are the associated saliency maps generated from the single-modal input which are sub-optimal. (c) is the RGB-D overlay. The state-of-the-art model \cite{zhouiccvspnet} fails to reason about an accurate saliency map under inferior conditions, i.e., inaccurate depth measurement and misalignment between both modalities (g). To address this issue, we propose a robust RGB-D fusion to explicitly model the depth noise for saliency detection, yielding results (h) closer to the ground-truth mask (d).} 

\label{fig:intro}
\vspace{-5mm}
\end{figure}

Saliency detection aims to segment image contents that visually attract human attention the most. Existing RGB-based saliency detection methods \cite{liu2019simple,zhang2017learningsingle,wu2019cascaded,zhao2020depth} achieve promising results in generic settings. However, in cluttered and visually similar backgrounds, they often fail to perform accurate detection. Therefore, many recent works 
\cite{zhao2020depth,piao2020a2dele,ji2020accurate,fan2019rethinkingd3} exploit image depths as additional geometric cues, in the form of RGB-D inputs,  to improve the saliency detection performance in difficult scenarios.

Given accurate and well-aligned depths, existing RGB-D methods perform well even in difficult scenarios. Unfortunately, this is not often the case in practice.
Sometimes, only low-quality depths can be acquired, depending upon the scene and the source of the depths. For example, depths from multi-view stereo cameras are often noisy~\cite{xu2021digging,cheng2020hierarchical} and asynchronous depth cameras are spatially  misaligned~\cite{paudel20192d}, as shown in Figure~\ref{fig:intro}. Other environmental factors such as object distance, texture, or even lighting conditions during the acquisition can also degrade the depth quality~\cite{fan2019rethinkingd3, modality2021wu, ji2021calibrated, Zhang2021DFMNet}. Therefore, a method that can still exploit the geometric cues, while being robust to the depth quality discrepancy is highly desired.

We observe that most existing methods perform unsatisfactorily on datasets with low-quality depths. This is primarily because of the commonly used fusion technique  \cite{fan2020bbs,pang2020hierarchical,CDINet,liu2021TriTrans,zhouiccvspnet,wu2021mobilesal} that merges the parallel streams of RGB and depth with equal importance while being agnostic to misalignment. Less accurate depths are expected to play a smaller role than their counterpart. On the other hand, the possibility of misalignment between RGB and depth needs to be considered during the fusion process.

 In this work, we propose a robust RGB-D fusion method that addresses the aforementioned problems of inaccurate and misaligned depths. The proposed method uses a layer-wise attention (LWA) mechanism to enable the depth quality aware fusion of RGB and depth features. Our LWA attention learns the trade-off between early and late fusions, depending upon the provided depth quality. More precisely, LWA encourages the early fusion of the depth features for high-quality depth inputs, and vice versa. Such fusion avoids the negative influence of the spurious depths while being opportunistic when high-quality depths are provided. In other words, the good-quality depth should play an important role in early layers thanks to its rich and exploitable low-level geometric cues, while low-quality depth should be more activated at semantic levels.
 
 To address the problem of misaligned depths, we introduce the trident spatial attention (TSA) that aggregates features from a wider spatial context. The introduced TSA is used to replace vanilla spatial attention, enabling the aligned aggregation of the misaligned features. In particular, our TSA requires only minor additional parameters and computation, while being sufficient to address the problem of misalignment. Note that the misalignment problem often exists only locally therefore the global context (at the cost of additional computation) may not be necessary. Such an example is shown in Figure~\ref{fig:intro}(c). We improve the vanilla spatial attention with different scales of receptive fields, yielding a simple yet efficient manner to replace the pixel-wise correspondence with region-wise correlation. Finally, the new spatial attention is adaptively merged with channel attention to form our hybrid fusion module. 
 
 In summary, our major contributions are listed below:

\begin{itemize}
\setlength{\itemsep}{1pt}
\setlength{\parsep}{0pt}
\setlength{\parskip}{0pt}
\item We study the problem of RGB-D fusion in a real-world setting, highlighting two major issues, inaccurate and misaligned depths, for accurate saliency detection.
    \item We introduce a novel layer-wise attention (LWA) to automatically adjust the depth contribution through different layers and to learn the best trade-off between early and late fusion with respect to depth quality. 
    \item We design a trident spatial attention (TSA) to better leverage the misaligned depth information by aggregating the features from a wider spatial context.
    
    \item Extensive comparisons on five benchmark datasets validate that our fusion performs consistently better than state-of-the-art alternatives while being very efficient.
\end{itemize}
%
\section{Related Work}
There are extensive surveys of salient object detection ~\cite{deepsodsurvey,borji2019sodsurvey,zhou2021rgbdsurvey} and on attention modules~\cite{tay2020efficient, khan2021transformers} in the literature. In the following, we briefly review related works. 

\begin{figure*}[t]
\centering
\includegraphics[width=0.95\linewidth,keepaspectratio]{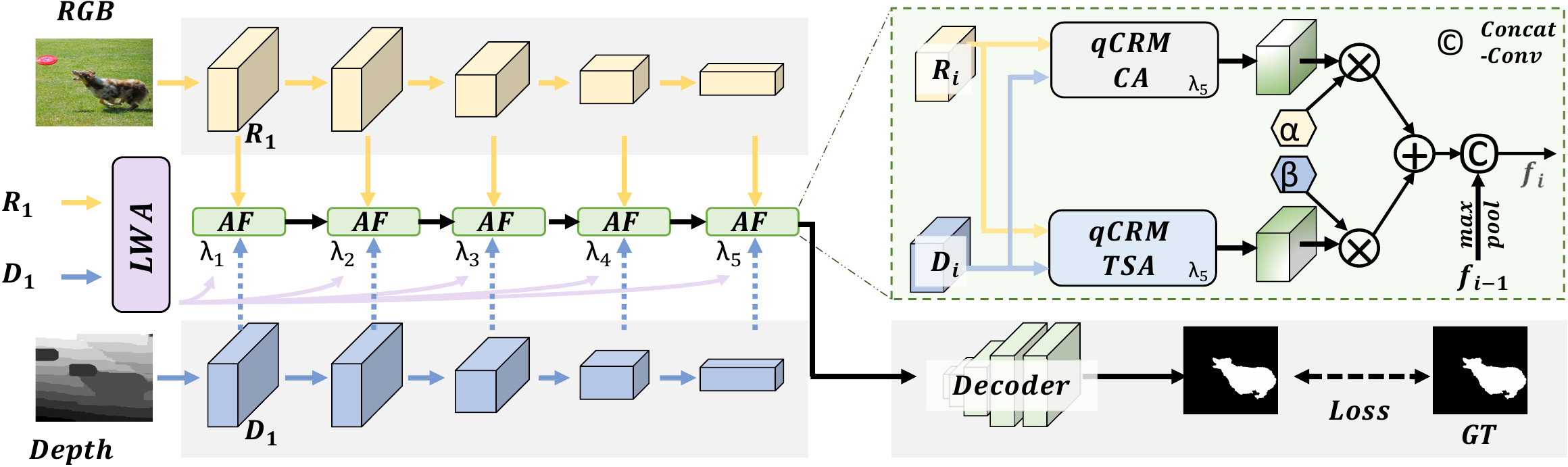}
\caption{\textbf{Architecture}. Our proposed network consists of a layer-wise attention (LWA, see Section~\ref{dqi}) and an adaptive Attention Fusion (AF, see Section~\ref{af}). LWA aims to find the best trade-off for early and late fusion depending on the depth quality, while AF leverages cross-modal cues to compute the shared representation with channel attention and improved spatial attention (TSA). CRM is from \cite{ji2021calibrated}.}
\label{fig:model}
\vspace{-2mm}
\end{figure*}

\noindent\textbf{RGB-D Fusion for Saliency Detection:} In the literature, we can divide current models into two types of architectures: single-stream and multi-stream schemes. The main difference is in the number of encoders. Single-stream networks are commonly lighter compared to multi-stream works. In~\cite{fu2020jldcf, zhao2020single}, the authors proposed to concatenate RGB-D images from the input side and then feed them into a single encoder-decoder architecture. From another perspective,~\cite{piao2020a2dele} introduces a depth distiller to enable cross-modal knowledge distillation, leading to a lightweight inference strategy with RGB-only input. Other works~\cite{zhao2019contrast, ji2020accurate, sun2021deep} propose to directly integrate low-level geometric cues in the RGB stream to strengthen the RGB features. Despite the proven result in previous works, single-stream models fail to explicitly analyze cross-modal correlation in complex scenarios, which is the main performance bottleneck.

Recently, multi-stream architectures have drawn increasing research interests. Several works~\cite{pang2020hierarchical, fan2020bbs, modality2021wu, zhouiccvspnet, cascaded_cmi, liu2021TriTrans, fang2022grouptransnet} propose to explicitly model RGB and depth cues through two parallel encoders and then aggregate multi-modal features through multi-scale fusion schemes, leading to better performance compared to their counterpart. In the literature, we can group existing works into three categories based on the fusion schemes: 1) depth-guided fusion, 2) discrepant fusion, and 3) multi-scale fusion. Depth enhanced fusion models~\cite{fan2020bbs, pang2020hierarchical, Zhang2021DFMNet} often adopts an asymmetric fusion scheme that the depth features are fused into RGB features at each level to improve boundary awareness. However, these models are sensitive to depth noise and the performance is significantly degraded when depth maps are under inferior conditions. Other works~\cite{pang2020hierarchical, zhang2020select, modality2021wu, CDINet, ji2021calibrated} propose to merge multi-modal cues through a discrepant design. In
~\cite{CDINet}, the authors adopt different fusion designs for low-level and high-level features, i.e., RGB to calibrate depth in earlier layers and depth to calibrate RGB in deeper layers. \cite{modality2021wu} adopts lightweight spatial attention \cite{wu:zacn,wu2022depth} only at semantic level. \cite{zhang2020select, ji2021calibrated} only fuse features at semantic levels, i.e., outputs from the last three layers. Different from discrepant and asymmetric designs, a number of works~\cite{zhao2020depth, zhouiccvspnet, cascaded_cmi, fang2022grouptransnet} realize bi-directional cross-modal interaction at each scale of the neural network. This fusion design, also known as middle fusion, has shown plausible performance in saliency benchmarks. Nevertheless, we observe that most existing works treat RGB and depth equally to form the shared features, paying little attention to explicitly modeling the measurement bias and alignment issue. \cite{Zhang2021DFMNet} has introduced a weighting strategy to deal with the measurement bias. However, their weighting scheme assumes the perfect alignment between multi-modal features. Different from previous works, we estimate the depth quality index by leveraging contextualized awareness. We show through empirical comparison that our approach can better model the depth quality to adjust the contribution.

\noindent\textbf{Attention for Cross-Modal Interaction} Self-attention modules~\cite{vaswani2017attention, woo2018cbam, wang2018non, fu2019dual,  wang2020eca, liu2021Swin} have been proven to be efficient for visual tasks. Inspired by their success, several RGB-D saliency works~\cite{fan2020bbs, zhao2020depth, ji2021calibrated, liu2020learning, CDINet} leverage self-attention as an augmentation to better preserve, calibrate, and fuse multi-modal features. \cite{zhao2020depth, ji2021calibrated} explicitly leverages the attention along the channel direction to calibrate each modality. \cite{liu2020learning} introduces a mutual and non-local strategy to learn the spatial cues from one modality and apply it to the other. Several recent works~\cite{liu2021TriTrans, liu2021vst, fang2022grouptransnet} further explore the long-range dependencies with transformer attention~\cite{vaswani2017attention}.  

Despite the popularity of contextualized attention, we observe that these modules often require a significant computational cost. Therefore, fusion with transformer attention is often realized with a small resolution feature map, i.e., at deeper layers of encoders~\cite{liu2020learning, liu2021TriTrans, fang2022grouptransnet}. To benefit from the spatial cues at each stage, a number of works~\cite{fan2020bbs, CDINet, fang2022grouptransnet} adopt the hybrid models with vanilla spatial and channel attention from~\cite{woo2018cbam} to aggregate features at each stage. However, vanilla spatial attention is agnostic of feature misalignment. Moreover, these hybrids treat spatial and channel attention equally, failing to be adjusted with respect to the network depth. Unlike previous works, we propose a simple yet efficient trident spatial attention that can better model contextualized awareness than its counterpart. Furthermore, we integrate our spatial attention with channel attention in a parallel scheme, yielding a more robust fusion strategy with adaptive weights.
%
\section{Method}
%
Before introducing the details, we highlight our technical motivation for better understanding the novelty of the proposed technique. D3Net \cite{fan2019rethinkingd3} is one of the pioneering works that explicitly model both modality-specific and fused saliency maps. The fusion design is realized at the output/saliency level. Differently, in our network, the RGB-D features are merged at the encoder stage. Several recent works \cite{liao2020mmnet,li2021hierarchical} fuse RGB-D features during encoder with the help of spatial attention. \cite{liao2020mmnet} uses RGB cues to improve depth, while \cite{li2021hierarchical} is bi-directional. From another perspective, DFMNet \cite{Zhang2021DFMNet} learns a weight to adjust the depth contribution. However, these methods focus more on the depth quality, paying little attention to the misalignment, i.e., vanilla spatial attention \cite{woo2018cbam} for \cite{liao2020mmnet,li2021hierarchical} or pixel-wise add/mul for \cite{Zhang2021DFMNet}. Differently, we explicitly decouple the low-quality and misalignment. We first leverage global attention to purely analyze the depth quality, based on which we introduce layer-wise attention to learn the best trade-off between early and late fusion. We show in Table \ref{tab:3dvablation} that our method outperforms the concurrent \cite{Zhang2021DFMNet}. Moreover, we propose an improved version of spatial attention with enlarged receptive fields. Compared to vanilla spatial attention, we show in Figure \ref{fig:fusion} and Table \ref{tab:3dvablation} that our improved version can better leverage multi-scale cues to tackle feature misalignment and yield superior performance.

Figure~\ref{fig:model} presents the overall framework of our Robust Fusion network (RFNet). We first extract RGB and depth features through parallel encoders. Then, these features are gradually merged through our proposed fusion module with respect to the depth noise. Specifically, to tackle the inaccurate measurement bias, we propose layer-wise attention (LWA) to control the depth contribution. To deal with feature misalignment, we propose a hybrid attention fusion (AF) module with a trident spatial attention and an adaptively merged channel attention. Details of each component are presented in the following sections.

\subsection{Layer-Wise Attention}\label{dqi}

\begin{figure}[t]
\centering
\includegraphics[width=\linewidth,keepaspectratio]{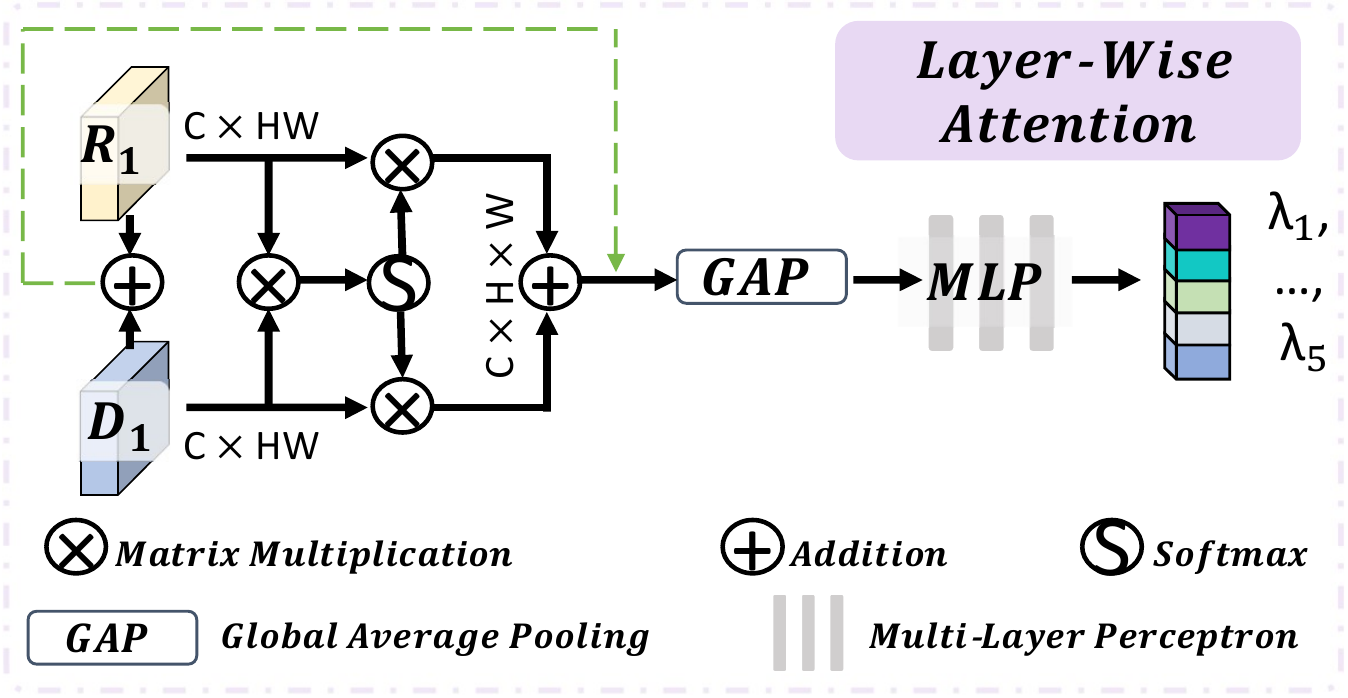}
\caption{\textbf{Layer-Wise Attention (LWA)}. It takes paired RGB and depth low-level features as input, i.e., features from first layer $R_1$ and $D_1$, and outputs confidence values $\lambda_i$ to adjust the depth contribution for the $i^{th}$ stage fusion. Specifically, we first leverage non-local attention to enable bi-directional interaction. Then, the cross-calibrated features are merged together and fed into an MLP to model the depth contribution. The dashed shortcut (\gr{green}) stands for the residual addition for reducing gradient vanishment.}
\label{fig:lwa}
\vspace{-2mm}
\end{figure}

We observe that there exist several depths with unsatisfactory quality as shown in Figure~\ref{fig:intro}. Inspired by this observation, we propose a depth quality indicator that aims to explicitly model the depth contribution. Our intuition is that while dealing with low-quality depth at early layers, the network should have a higher confidence value on the RGB feature instead of equally averaging the multi-modal cues.  

As depicted in Figure~\ref{fig:lwa}, our layer-wise attention takes outputs from the first encoder layer as input, i.e., $R_1 \in \mathbb{R}^{C \times H \times W}$ and $D_1 \in \mathbb{R}^{C \times H \times W}$. We argue that these features contain more heterogeneous and modality-specific cues compared to semantic-level features which are homogenized. With $R_1$ and $D_1$, we first compute the similarity between the two modalities. Instead of directly realizing the pixel-wise multiplication, we leverage the contextualized awareness to avoid the feature misalignment and focus on the measurement bias. Specifically, $R_1$ and $D_1$ are firstly fed into $Conv_{1\times 1}$ and flattened to form $R'_1 \in \mathbb{R}^{C \times HW}$ and $D'_1 \in \mathbb{R}^{C \times HW}$. These new features are then fed into the matrix multiplication. To normalize the obtained attention map, we further apply the softmax function to adjust the weight. Further, the normalized weight map is multiplied to flattened $R_1$ and $D_1$ to improve the cross-modal awareness. Finally, the retrieved RGB and depth attention maps are merged through addition.  Formally, the similarity matrix can be formulated as: 
\begin{equation}
   Attention(R'_1, D'_1) = softmax(\frac{R_1^{'} D_1^{'T}}{\sqrt{c}})(R'_1+ D'_1).
\end{equation}

Similar to self-attention works~\cite{vaswani2017attention, wang2018non}, we add a skip connection with early fused RGB-D features to stabilize the training procedure. Once we obtain the similarity matrix, we seek to explicitly quantify the depth measurement bias. Specifically, we first extract the feature vector with the help of global average pooling (GAP) and then feed it into a multi-level perceptron (MLP) to estimate the confidence values. We particularly estimate distinct values to explicitly guide feature fusion at different scales. The adaptive weight $\lambda \in \mathbb{R}^5$ can be formulated as:
\begin{equation}
    \lambda = MLP(GAP(Attention(R'_1, D'_1))).
\end{equation}

Finally, let $R_i$ and $D_i$ be the encoded RGB-D features from the $i^{th}$ layer. Instead of equally averaging both feature maps by $R_i + D_i$ which is agnostic of input depth quality, our proposed fusion by $R_i + \lambda_i D_i$ can better merge multi-modal features with context awareness.

At first glance, our attention map is similar to non-local attention~\cite{wang2018non} which has been applied in S2MA~\cite{liu2020learning} or to transformer attention~\cite{vaswani2017attention} which has been applied in TriTrans~\cite{liu2021TriTrans}. However, our method differs from previous works in two aspects, i.e., the purpose and the model size. Compared to S2MA which uses non-local attention for cross-modal calibration, our work aims to analyze the similarity between multi-modal features and assign a confidence value to the depth cues. Compared to TriTrans which adopts multi-head transformer attention to fuse features at the deepest layer, our design is significantly lighter with only one head and is applied to low-level features with higher resolution. The concurrent work DFMnet~\cite{Zhang2021DFMNet} adopts the Dice similarity coefficient~\cite{milletari2016v} to analyze the depth quality. However, it simply multiplies RGB and depth features with the pixel-wise association, paying little attention to explicitly model measurement bias and the misalignment in a separate manner.
\subsection{Adaptive Attention Fusion}\label{af}

Existing methods~\cite{fan2020bbs, zhao2020depth, ji2021calibrated, CDINet, fang2022grouptransnet} often adopt attention modules, i.e.,  spatial attention (SA) and channel attention (CA), to enable cross-modal interaction, with few methods pay attention to inherent feature misalignment. While by design CA is more robust to this issue due to the squeezed spatial resolution, the vanilla SA has more difficulties dealing with this inferior condition since it assumes a perfect alignment between different modalities. To address this dilemma, we propose to improve the current SA with enlarged global awareness, yielding a simple yet efficient manner to replace the pixel-wise alignment with region-wise correlation. Furthermore, current works simply apply CA and SA one by another \cite{fan2020bbs, fang2022grouptransnet} or equally average them to form the output \cite{CDINet}. These works are agnostic to the network depth that SA and CA still contribute equally at each stage. Previous work~\cite{pan2021integration} has shown that layers with different depths will pay attention to different contexts. Therefore, we seek to introduce an adaptive fusion strategy with learnable weights to automatically adjust the contribution of each attention at different levels.

\begin{figure}[t]
\centering
\includegraphics[width=\linewidth,keepaspectratio]{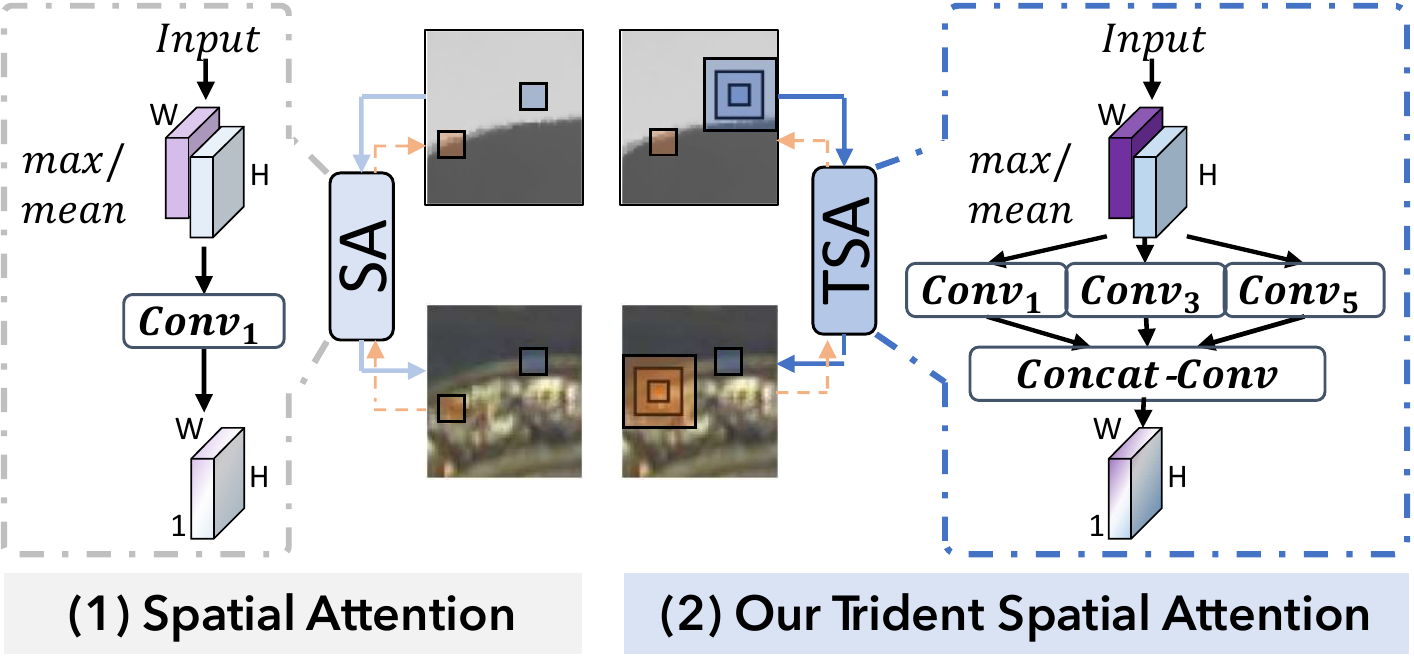}
\caption{\textbf{Motivation of our attention fusion}: (1) Vanilla spatial attention~\cite{woo2018cbam, fan2020bbs, CDINet} which is not suitable for cross-modal interaction due to feature misalignment. (2) We propose a trident spatial attention (TSA) with dilated receptive field to better leverage contextualized awareness. Better to zoom in.}
\label{fig:fusion}
\vspace{-2mm}
\end{figure}

Formally, let an input feature map $f \in \mathbb{R}^{C \times H \times W}$. The vanilla SA firstly squeezes the channel dimension with average and max pooling across the channel, denoted as $CAP(\cdot)$ and $CMP(\cdot)$, respectively, to obtain the spatial map $f' \in \mathbb{R}^{2\times H \times W}$ . Then, from $f'$, SA learns a 2-D weight map $SA \in \mathbb{R}^{1\times H \times W}$:
\begin{equation}
\begin{split}
&f' = Concat(CAP(f), CMP(f)); \\
&SA(f) = \sigma(Conv_1(f'))),
\end{split}
\end{equation}
where $\sigma(\cdot)$ is the Sigmoid activation, $Conv_1$ stands for the convolution with dilation 1. To improve global awareness, as shown in Figure \ref{fig:fusion} we replace the current convolution with trident branches where each branch focuses on learning features with different scales. Our proposed trident spatial attention can be formulated as:
\begin{equation}
\begin{split}
TSA(f) = \sigma(Concat( & Conv_1(f') \\
& Conv_3(f') \\
& Conv_5(f')).
\end{split}
\end{equation}
where $Conv_1, Conv_3, Conv_5$ stand for convolutions with different dilation values.

To attentively aggregate multi-modal features, we follow the pipeline of the Cross-Reference Module (CRM) as suggested in DCF~\cite{ji2021calibrated}. Formally, let $R$ and $D$ the paired RGB-D input for the fusion module, we first compute the modality-specific channel $CA_r$ and $CA_d$, as well as the shared channel attention $CA_f$ as follow:
\begin{equation}
\begin{split}
& CA_r = CA(R); \quad CA_d = CA(D); \\
& CA_f = norm(max(CA_r, CA_d)); \\
\end{split}
\end{equation}

The vanilla CRM benefits from channel attention to realize the self- and cross-calibration before the feature fusion. We have:
\begin{equation}
\begin{split}
CRM(R,D) = Concat( & CA_f \otimes CA_r \otimes R;  \\
& CA_f \otimes CA_d \otimes D) ; \\
\end{split}
\end{equation}

We refer readers to the original paper~\cite{ji2021calibrated} for more details on the cross-modal interaction. In our application, we replace the final concatenation with adaptive addition with respect to depth quality and form our $qCRM^{CA}$ as follows:

\begin{equation}
\begin{split}
qCRM^{CA}(R,D) = & CA_f \otimes CA_r \otimes R +  \\
& \lambda \cdot CA_f \otimes CA_d \otimes D ; \\
\end{split}
\end{equation}

Moreover, we additionally design another branch where the CA is replaced by our proposed TSA. This new branch is termed as $qCRM^{TSA}$. We further learn two scalar values $\alpha$ and $\beta$ to adaptively weight $CRM^{TSA}$ with the original branch $CRM$ with channel attention. Our adaptive fusion (AF) can be formulated as:
\begin{equation}
AF(R, D) = \alpha \cdot qCRM^{CA}(R,D)  + \beta \cdot qCRM^{TSA}(R,D)
\label{eq:fusion}
\end{equation}
Finally, we merge the previous level output $f_{i-1}$, if any, and the current AF output with concatenation-convolution. To deal with the resolution, we apply max-pooling on $f_{i-1}$ to preserve the most informative hierarchical features.

\subsection{Architecture}

In this paper, we propose a novel fusion design that can be easily adapted to any existing architecture. To compete with the state-of-the-art performance, we choose Res2Net~\cite{gao2019res2net} as our backbone to extract features. Our decoder is the same as SPNet~\cite{zhouiccvspnet}. Specifically, it consists of five-level RFB blocks~\cite{liu2018rfb}. Each block is skipped and connected with the fused encoded features. However, different from SPNet with a triple decoder to explicitly both modality-specific and shared features, we only maintain one decoder to decode our efficiently fused features. Our network is supervised by conventional IoU and BCE losses.
%
\section{Experimental Validation}
%
\begin{table*}[t]
\scriptsize
\setlength\tabcolsep{0.8pt}
\setlength\extrarowheight{2pt}
\begin{center}
\caption{Quantitative comparison with different fusion designs. We replace our fusion module with five SOTA fusion modules and retrain the new networks with the same training setting. $\uparrow$ ($\downarrow$) denotes that the higher (lower) is better.}
\label{tab:3dvfusion}
\begin{tabular*}{\textwidth}{@{\extracolsep{\fill}}*{22}{l}}
\hline

\hline

\hline
Dataset &Size$\downarrow$ &\multicolumn{4}{c}{DES} & \multicolumn{4}{c}{NLPR} & \multicolumn{4}{c}{NJU2K} & \multicolumn{4}{c}{STERE} & \multicolumn{4}{c}{SIP} \\
\cline{3-6} \cline{7-10} \cline{11-14} \cline{15-18} \cline{19-22} 

Metric & $(\Delta$ Mb)  &
        $M\downarrow$ & $F\uparrow$ &  $S\uparrow $ & $E\uparrow$ &
        $M\downarrow$ & $F\uparrow$ &  $S\uparrow $ & $E\uparrow$ &
        $M\downarrow$ & $F\uparrow$ &  $S\uparrow $ & $E\uparrow$ &
        $M\downarrow$ & $F\uparrow$ &  $S\uparrow $ & $E\uparrow$ &
        $M\downarrow$ & $F\uparrow$ &  $S\uparrow $ & $E\uparrow$ \\
\hline
BBS \cite{fan2020bbs}    & 460(+95)       &   .015&   \textbf{.946}&   \textbf{.941}&   .976
                                     &   .023&   .920&   .923&   .953
                                     &   .035&   .924&   .915&   .944
                                     &   .040&   .913&   .902&   .935
                                     &   .053&   .904&   .877&   .912\\
                                
DFM\cite{Zhang2021DFMNet}    &495(+130)   &   .015&   \textbf{.946}&   \textbf{.941}&   .974
                                     &   .022&   .919&   .926&   .956
                                     &   .034&   .922&   .917&   .943
                                     &   .041&   .909&   .902&   .933 
                                     &   .049&   .909&   .885&   .919\\	

CDI\cite{CDINet}    &520(+155)   &   .023&   .919&   .914&   .950
                                     &   .024&   .918&   .921&   .952
                                     &   .035&   .927&   .915&   .944
                                     &   .036&   .918&   .910&   .941 
                                     &   .055&   .900&   .870&   .911\\

DCF\cite{ji2021calibrated}  &336(\textbf{-29})    &    .015&   .944&   .938&   .976
                                     &    \textbf{.020}&   .927&   \textbf{.931}&   .960
                                     &    .030&   .930&   .924&   .949
                                     &    .038&   .913&   .904&   .937 
                                     &    .044&   .913&   .891&   .928\\
           
SPNet\cite{zhouiccvspnet}  & 593(+228)  
                                     &  .016&   .944&   .936&   .973 
                                     &  .022&   .924&   .925&   .956
                                     &  .032&   .928&   .919&   .945 
                                     &  .038&   .913&   .904&   .938 
                                     &  .048&   .907&   .884&   .921\\

MobSal\cite{wu2021mobilesal}&723(+358)   &   \textbf{.015}&  .945&   .940&   .976
                                   &   .024&   .924&   .923&   .955
                                    &   .033&   .926&   .915&   .945
                                    &   .038&   .913&   .902&   .937 
                                    &   \textbf{.042}&   .915& .892&   .930\\
\hline
Ours &365 
&\textbf{.015}&\textbf{.946}&\textbf{.941}&\textbf{.977}
&\textbf{.020}&\textbf{.932}&\textbf{.931}&\textbf{.962}
&\textbf{.029}&\textbf{.936}&\textbf{.926}&\textbf{.951}
&\textbf{.035}&\textbf{.921}&\textbf{.911}&\textbf{.944} 
&\textbf{.042}&\textbf{.916}&\textbf{.893}&\textbf{.931}\\
\hline 

\hline 

\hline
\end{tabular*}
\end{center}
\vspace{-5mm}
\end{table*}

\subsection{Datasets, Metrics and Training Settings}
We follow previous works~\cite{fan2020bbs, modality2021wu, ji2021calibrated, zhouiccvspnet} and train our model on the conventional training set which contains 1,485 samples from the NJU2K-train~\cite{ju2014depth} and 700 samples from the NLPR-train~\cite{peng2014rgbd}. 
For testing benchmarks,  we observe that the depth quality within each dataset varies, which is mainly due to acquisition methods. Specifically, DES \cite{cheng2014depth} contains 135 images of indoor scenes captured by a Kinect camera. SIP \cite{fan2019rethinkingd3} provides a human dataset that contains 929 images captured by a mobile device. Therefore, these two datasets can be considered moderate with less noisy depths. 

The remaining NLPR-test~\cite{peng2014rgbd}, NJU2K-test~\cite{ju2014depth} and STERE~\cite{niu2012leveraging} datasets are more challenging. NLPR-test~\cite{peng2014rgbd} contains 300 natural images which are captured by a Kinect sensor. However, the images are obtained under different illumination conditions. NJU2K-test \cite{ju2014depth} contains 500 stereo image pairs from different sources such as the Internet and 3D movies. A number of depth maps are estimated through the optical flow method~\cite{sun2010secrets}. STERE~\cite{niu2012leveraging} contains 1,000 stereoscopic images where the depths are estimated with SIFT flow method \cite{liu2010sift}. Due to the measurement or estimation error, these datasets contain more noisy depths. Therefore, to purely analyze the performance under different conditions, we additionally report the average metric (AvgMetric) for datasets with good quality depths and for datasets with more challenging depths. 

To quantify the performance of our methods, we use conventional saliency metrics such as Mean Absolute Error, F-measure, S-measure, and E-measure. More details can be found in the supplementary material. 


Our method is based on the Pytorch framework and is learned with a V100 GPU. The encoder is initialized with the pre-trained weights. For the 1-channel depth input, we replace the first convolution of backbone to feet with the depth size.  The learning rate is initialized to 1$e-$4 which is further divided by 10 every 60 epochs. We fix and resize the input RGB-D resolution to 352$\times$352. During training, we adopt random flipping, rotating, and border clipping for data augmentation. The total training time takes around 5 hours with batch size 10 and epoch 100.

\subsection{Comparison with SOTA fusion alternatives}\label{fusion}
We observe that existing works adopt different architectures, i.e., choice of backbones, design of decoder, supervision, training settings, etc. For example, light models \cite{Zhang2021DFMNet, wu2021mobilesal} always choose MobileNet \cite{sandler2018mobilenetv2} to extract features. Several works~\cite{pang2020hierarchical, zhao2020single, piao2020a2dele, jin2021cdnet} are based on VGG encoders \cite{simonyan2014vgg} , while another group of models~\cite{Zhang2021DFMNet, fang2022grouptransnet} takes ResNet \cite{He2016Residual} as encoders. Recent works~\cite{zhouiccvspnet, liu2021TriTrans} are based on more powerful backbones such as Res2Net \cite{gao2019res2net} and ViT \cite{vit}. The choice of backbone will undoubtedly impact the final performance.  Furthermore, the design of the decoder varies from one work to another. Several works are based on DenseASPP~\cite{yang2018denseaspp}, while others are based on RFB~\cite{liu2018rfb}. Under the consideration of a fair comparison, we re-implement six SOTA fusion works under the same architecture. Specifically, we choose the same backbone, same decoder, loss, and same training settings as ours. The only difference between one model to another is in the fusion module. We refer readers to previous sections for more experimental details. Note that several fusion designs~\cite{wu2021mobilesal, ji2021calibrated} were initially applied only to certain layers. To fairly and purely analyze the fusion performance, we implement all the fusion modules at each scale as ours.

Table~\ref{tab:3dvfusion} illustrates the quantitative comparison. We also report the model size of each embedded fusion module. $\Delta Size$ stands for the difference in model size compared to ours. It can be seen that our fusion strategy yields significantly better results compared to our counterparts. Compared to the lightest DCF fusion which only applies channel attention during feature fusion, we add additional spatial attention, yielding a slightly heavier model size (+29 Mb) but favorably improving the performance. Elsewise, our model size is significantly lighter compared to other counterparts, validating the effectiveness of our proposed fusion module.

\begin{table*}[t]
\scriptsize
\setlength\tabcolsep{0pt}
\setlength\extrarowheight{0.05pt}
\begin{center}
\caption{Quantitative comparison with state-of-the-art models. $\uparrow$ ($\downarrow$) denotes that the higher (lower) is better. The best and second best are highlighted in \textbf{bold} and \underline{underline}, respectively. We further report the average metric (AvgMetric) for datasets with more challenging depths and with less noisy depths.}
\label{tab:3dvquant}
\begin{tabular*}{\textwidth}{@{\extracolsep{\fill}}*{2}{l}|@{\extracolsep{\fill}}*{12}{l}@{\extracolsep{\fill}}*{4}{l}|@{\extracolsep{\fill}}*{14}{l}}
\hline

\hline

\hline
 & & \multicolumn{16}{c|}{Benchmarks with challenging depth} & \multicolumn{12}{c}{Benchmarks with less noisy depth}\\

Dataset & Size$\downarrow$ & \multicolumn{4}{c}{NLPR} & \multicolumn{4}{c}{NJU2K} & \multicolumn{4}{c}{STERE} &  \multicolumn{4}{c|}{\textbf{AvgMetric}}  &\multicolumn{4}{c}{DES} &
\multicolumn{4}{c}{SIP} & \multicolumn{4}{c}{\textbf{AvgMetric}}\\ 
\cline{3-6} \cline{7-10} \cline{11-14} \cline{15-18} \cline{19-22} \cline{23-26}
\cline{27-30}

Metric & (Mb) & $M\downarrow$ & 
$F\uparrow $ & $S\uparrow$ & $E\uparrow$ & $M\downarrow$ & $F\uparrow $ & $S\uparrow$ & $E\uparrow$ & $M\downarrow$ &
$F\uparrow $ & $S\uparrow$ & $E\uparrow$ & $M\downarrow$ &
$F\uparrow $ & $S\uparrow$ & $E\uparrow$ & $M\downarrow$ & $F\uparrow $ & $S\uparrow$ & $E\uparrow$ & $M\downarrow$ & $F\uparrow $ & $S\uparrow$ & $E\uparrow$ & $M\downarrow$ & 
$F\uparrow $ & $S\uparrow$ & $E\uparrow$\\

\hline

$CPFP_{19}$ \cite{zhao2019cpfp}                   &278 
                                
                                & .036 & .867 & .888 &.932 
                                & .053 & .877 & .878 &.923 
                                & .051 & .874 & .879 &.925
                                & .049 & .873 & .880 &.925
                                & .038 & .846 & .872 &.923 
                                & .064 & .851 & .850 &.903
                                & .060 & .850 & .852 &.905\\ 

$DMRA_{19}$  \cite{piao2019dmra}        &238  
                                & .031 & .879 & .899 &.947 
                                & .051 & .886 & .886 &.927 
                                & .047 & .886 & .886 &.938
                                & .045 & .884 & .888 &.936
                                & .030 & .888 & .900 &.943 
                                & .085 & .821 & .806 &.875
                                & .078 & .829 & .817 &.883\\ 

$A2dele_{20}$  \cite{piao2020a2dele}    & \textbf{116}   
                                & .029 & .882 & .898 &.944 
                                & .051 & .874 & .871 &.916 
                                & .044 & .879 & .878 &.928 
                                & .043 & .878 & .879 &.927	
                                & .029 & .872 & .886 &.920 
                                & .070 & .833 & .828 &.889
                                & .060 & .850 & .852 &.905\\ 

$JLDCF_{20}$  \cite{fu2020jldcf}         &548  
                                & .022 & .916 & .925 & \underline{.962}
                                & .043 & .903 & .903 & .944    
                                & .042 & .901 & .905 & .946
                                & .038 & .904 & .907 & .948
                                & .022 & .919 & .929 &.968
                                & .051 & .885 & .879 &.923 
                                & .047 & .889 & .885 &.928\\ 

$CMMS_{20}$  \cite{li2020rgb}       & 546   
                                & .027 & .896 & .915 &.949 
                                & .044 & .897 & .900 &.936 
                                & .043 & .893 & .895 &.939 
                                & .040 & .894 & .899 &.939
                                 & .018 & .930 & .937 &.976 
                                & .058 & .877 & .872 &.911
                                & .052 & .883 &  .880 & .918\\ 

$CoNet_{20}$   \cite{ji2020accurate}          &162        
                                & .031 & .887 & .908 &.945 
                                & .046 & .893 & .895 &.937 
                                & .040 & .905 & .908 &\textbf{.949} 
                                & .040 & .898 & .904 &.945
                                & .028 & .896 & .909 &.945
                                & .063 & .867 & .858 &.913
                                & .058 & .870 & .864 &.917\\ 

$DANet_{20}$   \cite{zhao2020single}       & \underline{128}        
                                & .028 & .916 & .915 &.953 
                                & .045 & .910 & .899 &.935 
                                & .043 & .892 & .901 &.937 
                                & .041 & .901 & .902 &.939
                                & .023 & .928 & .924 &.968 
                                & .054 & .892 & .875 &.918
                                & .050 & .896 & .881 &.924\\ 

$DASNet_{20}$   \cite{zhao2020depth}        & -       
                                & \underline{.021} & .929 & \underline{.929} &- 
                                & .042 & .911 & .902 &- 
                                & .037 & .915 & \underline{.910}  & -
                                & .035 & .916 & .910  & -
                                 & .023 & .928 & .908 &- 
                                &- &- & - & -
                                &- &- & - & - \\ 

$HDFNet_{20}$  \cite{pang2020hierarchical}       &308          
                                & .031 & .839 & .898 &.942 
                                & .051 & .847 & .885 &.920 
                                & .039 & .863 & .906 &.937
                                & .041 & .854 & .898 &.933
                                & .030 & .843 & .899 &.944
                                & .050 & \underline{.904} & .878 &.920
                                & .047 & .896 & .880 &.923	\\ 

$BBSNet_{20}$    \cite{fan2020bbs}      &200           
                                & .023 & .918 & \underline{.930} &.961 
                                & .035 & .920 & .921 &.949 
                                & .041 & .903 & .908 &.942 
                                & .036 & .910 & .915 &.947
                                & .021 & .927 & .933 &.966 & - & - & - &- & - & - & - &- 	\\ 

$DCF_{21}$   \cite{ji2021calibrated}           &435      
                                & \underline{.021} & .891 & - &.957 
                                & .035 & .902 & - &.924 
                                & .039 & .885 & - &.927 
                                & .034 & .890 & - &.931
                                & - & - & - &- 
                                & .051 & .875 & - &.920 
                                 & - & - & - &-  \\
                                
$D3Net_{21}$  \cite{fan2019rethinkingd3}          &518
                                & .030 & .897 & .912 &.953 
                                & .041 & .900 & .900 &.950 
                                & .046 & .891 & .899 &.938 
                                & .041 & .894 & .901 &.943
                                & .031 & .885 & .898 &.946 
                                & .063 & .861 & .860 &.909
                                & .058 & .864 & .864 &.913\\ 

$DSA2F_{21}$   \cite{sun2021deep}           & -     
                                & .024 & .897 & .918 &.950 
                                & .039 & .901 & .903 &.923 
                                & .036 & .898 & .904 &.933
                                & .034 & .898 & .906 &.933
                                & .021 & .896 & .920 &.962 
                                & - & - & - &- 
                                & - & - & - &-\\ 


$TriTrans_{21}$  \cite{liu2021TriTrans}   
&927 
                                & .020 & .923 & .928 &.960 
                                & .030 & .926 & .920 &.925 
                                & \textbf{.033} & .911 & .908 &.927 
                                & .030 & .917 & .914 &.931
                                & .014 & .940 & .943 &.981 
                                & .043 & .898 & .886 &.924
                                & .039 & - & .893 &\underline{.931} \\ 

$CDINet_{21}$  \cite{CDINet}       &217 
                                & .024 & .916 & .927 &- 
                                & .035 & .922 & .919 &- 
                                & .041 & .903 & .906 &-
                                & .036 & .910 & .913 &-
                                & - & - & - &- 
                                & - & - & - &- 
                                & - & - & - &- \\ 
                                

$SPNet_{21}$\cite{zhouiccvspnet}
& 702
                                & \underline{.021} & .925 & .927 &.959 
                                & \textbf{.028} & \underline{.935} & \underline{.925} &\textbf{.954} 
                                & .037 & \underline{.915} & .907 &\underline{.944}
                                & \underline{.031} & \underline{.922} & \underline{.915} &\textbf{.949}
                                & \textbf{.014} & \textbf{.950} & \textbf{.945} &\textbf{.980} 
                                & .043 & \textbf{.916} & \textbf{.894} &\underline{.930}
                                & \underline{.039} & \textbf{.920} & \textbf{.900} &\textbf{.936}\\ 


%

\hline
RFNet (ours)    
&  364 
                                & \textbf{.020}& \textbf{.932}  &\textbf{.931}&   \textbf{.962}
                                & \underline{.029}& \textbf{.936}&  \textbf{.926}&   \underline{.951}
                                & \underline{.035}& \textbf{.921}&  \textbf{.911}&   \underline{.944}
                                & \textbf{.030}&	\textbf{.927}&  \textbf{.918}&   \underline{.948}
                                & \underline{.015}& \underline{.946}&  \underline{.941}&   \underline{.977} 
                                & \textbf{.042}& \textbf{.916}&  \underline{.893}&   \textbf{.931}
                                & \textbf{.038}& \underline{.919}&  \underline{.899}&   \textbf{.936}
                               \\ 

\hline

\hline

\hline
\vspace{-8mm}

\end{tabular*}
\end{center}

\end{table*}

\subsection{Quantitative Comparison}

\begin{figure}[t]
\centering
\includegraphics[width=0.95\linewidth,keepaspectratio]{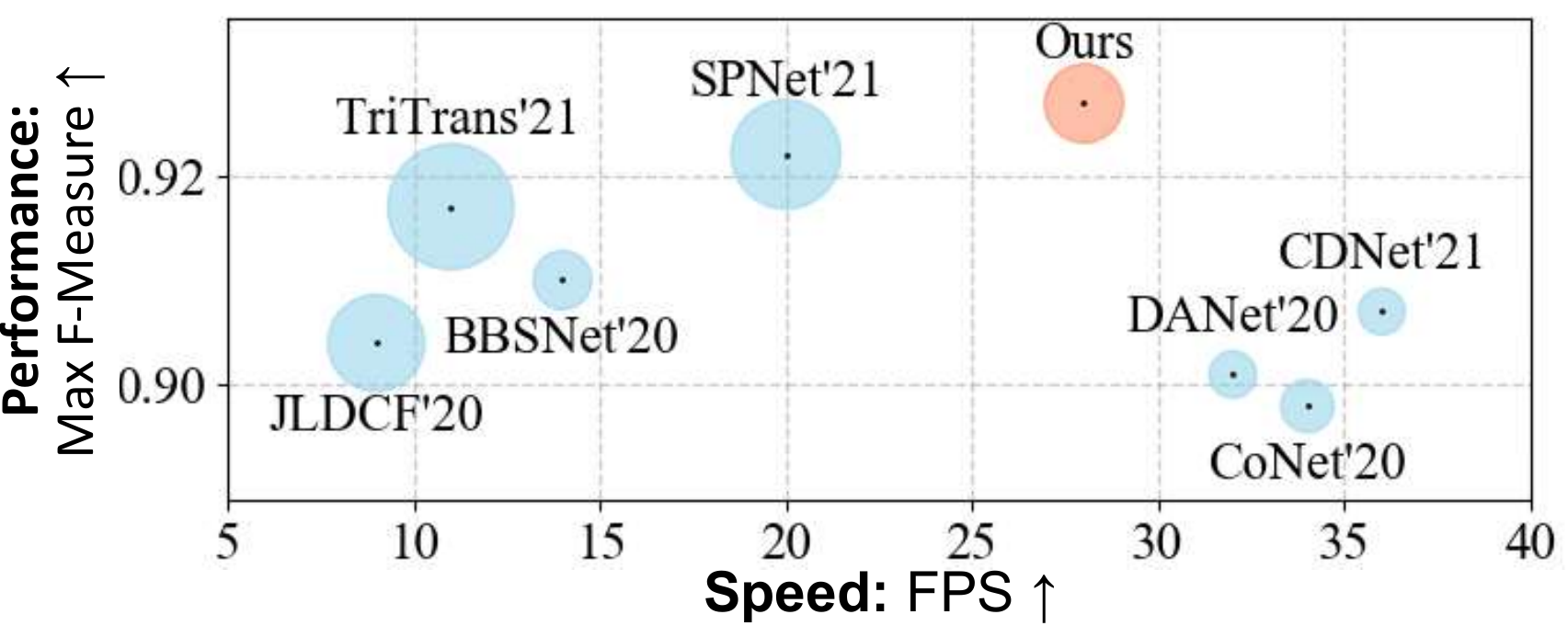}

\caption{Average \textbf{Performance, Speed, and Model Size} of different methods on challenging datasets (NLPR, NJUK, STERE). The circle size denotes the model size. Note that better models are shown in the upper right corner (i.e., with a larger F-measure and larger FPS). Our method finds the best trade-off of the three measures. Methods with higher speed perform inferior, making our method both efficient and accurate.}
\label{fig:plot}
\vspace{-3mm}
\end{figure}

\begin{figure*}[t]
\centering
\includegraphics[width=.95\linewidth,keepaspectratio]{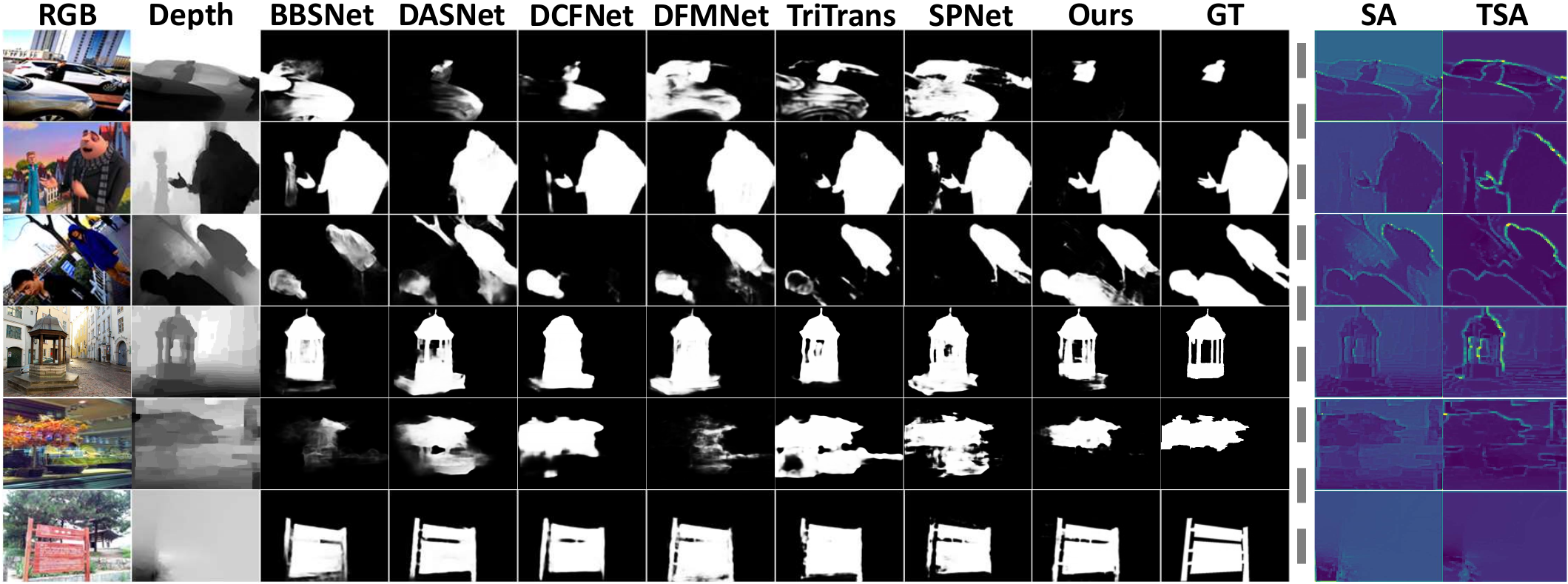}
\caption{\textbf{Qualitative comparison}. We also illustrate the depth features enhanced by vanilla SA and by our proposed TSA, respectively. Our work yields more boundary activation compared to the counterpart. Better to zoom in.}
\label{fig:quali}
\vspace{-4mm}
\end{figure*}

Table~\ref{tab:3dvquant} illustrates the quantitative comparison. For challenging datasets (NLPR, NJU2K, and STERE), our method performs favorably over the existing methods and sets a new state-of-the-art (SOTA) record, validating the superior robustness of our approach against depth bias. We further illustrate in Figure \ref{fig:plot} the trade-off between model efficiency and SOTA performances. Compared to the current SOTA TriTrans \cite{liu2021TriTrans} with 11 FPS and SPNet \cite{zhouiccvspnet} with 20 FPS, our network achieves superior performance with higher inference speed, i.e., 28 FPS. For other datasets with less depth noise (DES and SIP), we also achieve competitive performance with almost halved the model size compared to the current SOTA SPNet \cite{zhouiccvspnet}. Note that both SPNet and ours adopt Res2Net50 \cite{gao2019res2net} as the backbone. Thus, our performance can be contributed to our proposed fusion solely. 

\subsection{Qualitative Comparison}

Figure~\ref{fig:quali} presents the generated saliency maps of different methods on challenging cases such as single or multiple humans, clustered foreground-background, and low-quality depth. It can be seen that our methods consistently reason about saliency masks closer to the ground truth. We further illustrate the comparison between depth feature maps enhanced with our proposed spatial attention (TSA) and with the counterpart (TA). We can visualize that our attention is more sensitive to camera distances and can better segment object regions. This can be contributed to our trident branches with different scales.  Furthermore, our attention yields more activation on the boundary, facilitating the network to better leverage geometry for saliency detection. 

\begin{figure}[t]
\centering
\includegraphics[width=\linewidth,keepaspectratio]{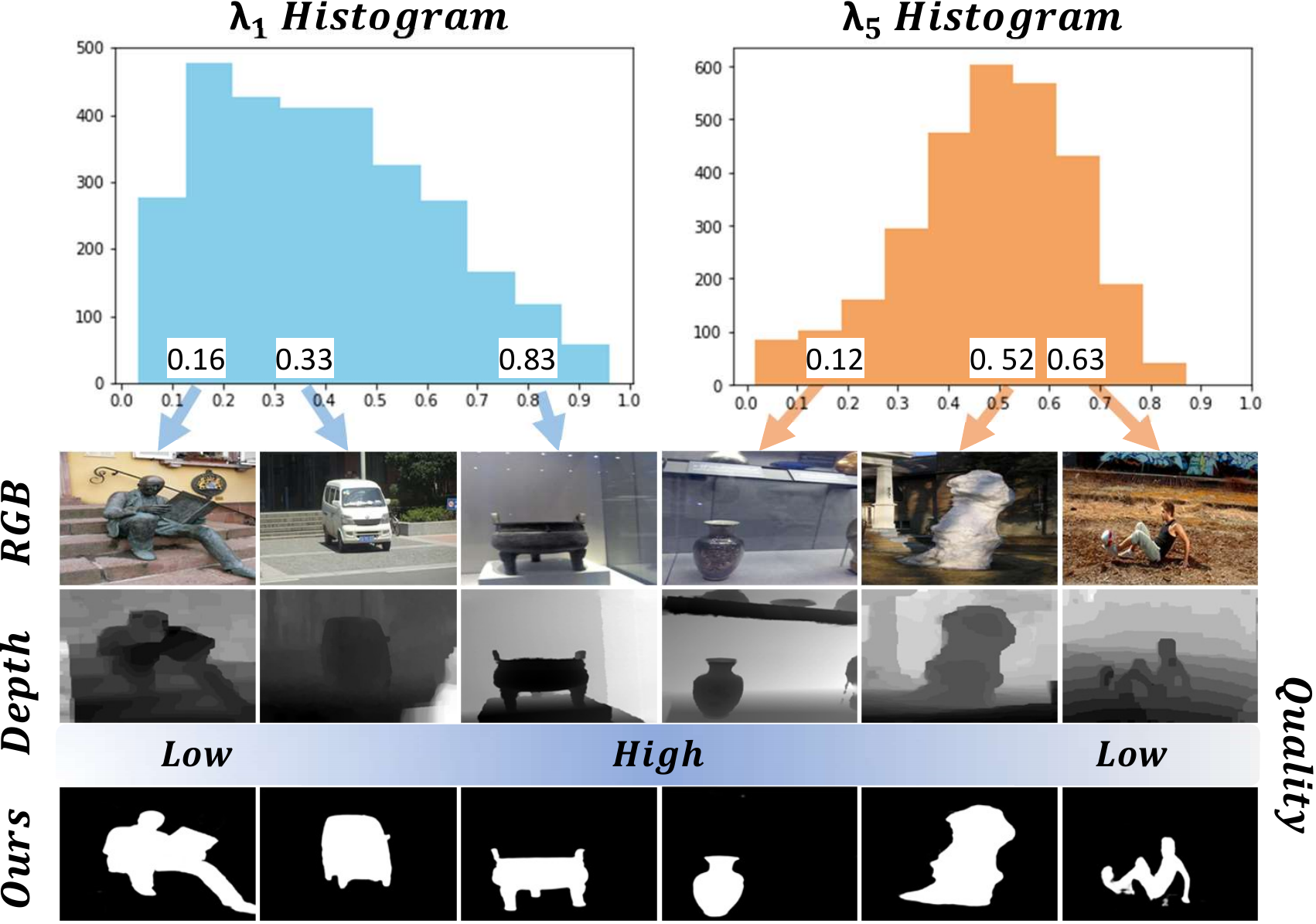}
\caption{\textbf{Trade-off between early and late fusion}. Our layer-wise attention can adaptively model the depth contribution during feature fusion. While are of low quality, we assign less weight to early fusion since the noisy geometric cues are difficult to be exploited. Meanwhile, we assign more weight to late fusion to leverage the multi-modal semantic cues for feature fusion.}
\label{fig:contribution}
\vspace{-0mm}
\end{figure}

Finally, we illustrate in Figure \ref{fig:contribution} the histogram for our layer-wise attention. We particularly choose $\lambda_1$ and $\lambda_5$ to facilitate the understanding of the trade-off between early and late fusion. We can observe that while depths are of low quality, our LWA assigns more weights for late fusion (with low $\lambda_1$ value and high $\lambda_5$ value). While depths are of good quality, our LWA assigns more weights for early fusion (with high $\lambda_1$ value and low $\lambda_5$ value). This observation is consistent with previous studies \cite{CDINet,pang2020hierarchical,ji2021calibrated,fang2022grouptransnet} with discrepant fusion. We hope our analysis of layer-wise attention can inspire future adaptive fusion works.

\begin{figure}[t]
\centering
\includegraphics[width=\linewidth,keepaspectratio]{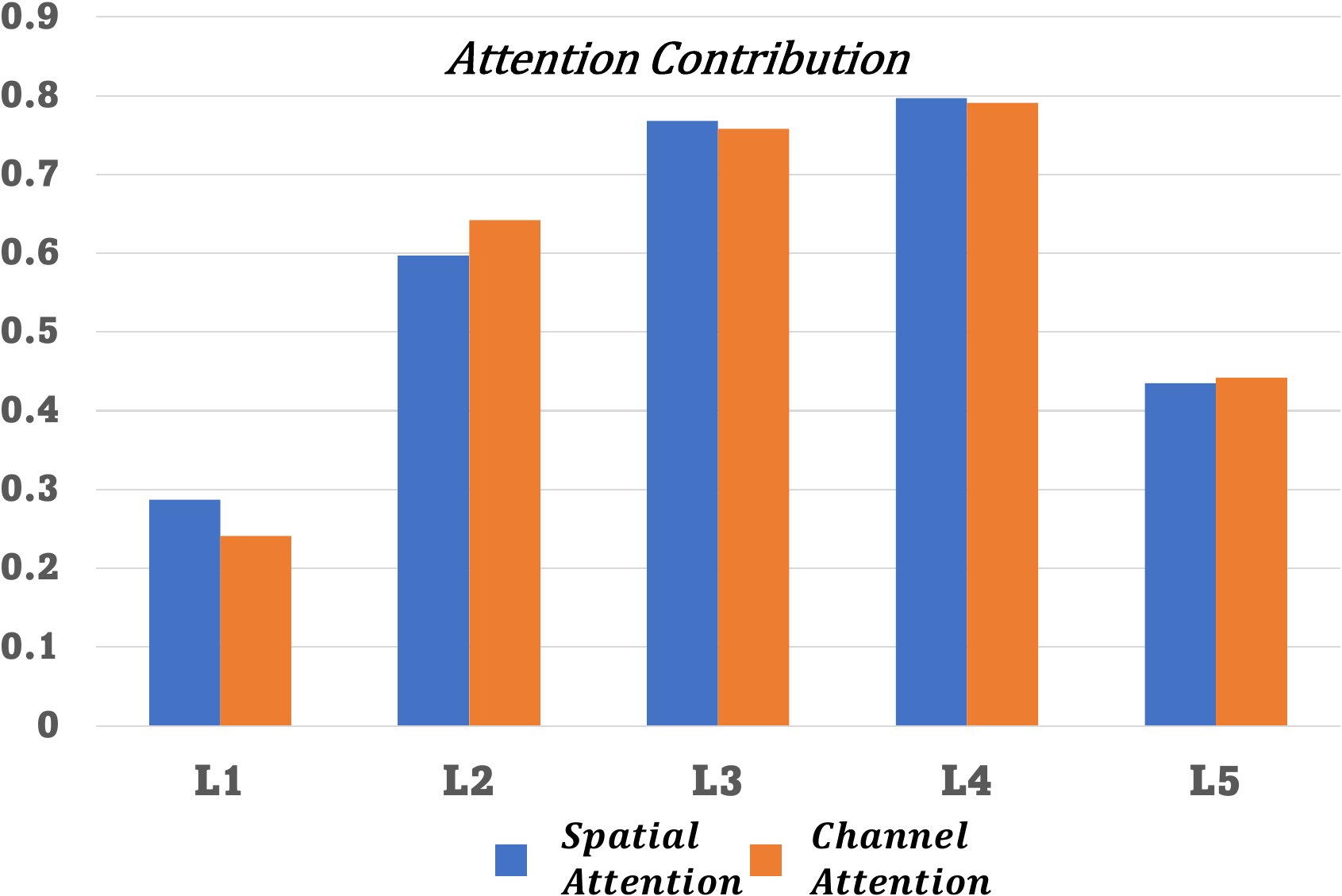}
\caption{\textbf{Attention contribution}. $L_1, ... L_5$ stands for the different layers. The attention fusion is described in Eq. \ref{eq:fusion}.}
\label{fig:adaptive}
\vspace{-2mm}
\end{figure}

\subsection{Distribution of Spatial and Channel Attention}

Since we propose an adaptive weighting strategy to merge our spatial attention (TSA) and channel attention (CA), we illustrate in Figure~\ref{fig:adaptive} the distribution of weights of each attention at different stages of the network. We can observe that TSA and CA contribute differently with respect to the network depth. At layer 1 ($L_1$), the network assigns more weights to TSA, which can be explained by the significant spatial resolution of the features. For deeper layers, TSA and CA tend to play a similar role at each stage to enhance the feature modeling with equal importance from both spatial and channel cues. However, we can observe that attention from different layers contributes differently to the final output, i.e., attention from the third layer ($L_3$) and the fourth layer ($L_4$) contribute more compared to the first two layers ($L_1-L_2$) and the last layer ($L_5$). The difference with respect to the network depth is also consistent with previous work \cite{pan2021integration} and to our layer-wise attention that shallow and deep layers play different roles for feature fusion.

\subsection{Ablation Study}
In this section, we conduct an ablation study to validate the effectiveness of each proposed component. The quantitative result of each combination can be found in Table~\ref{tab:3dvablation}. To analyze the effectiveness of our trident spatial attention (TSA), we replace ours with vanilla spatial attention~\cite{woo2018cbam} and observe a dropped performance. This is mainly due to the limited receptive field of vanilla attention that assumes a local correlation between different features. In contrast, our TSA can significantly improve performance by leveraging contextualized awareness. The boosted performance on the aforementioned datasets validates the design of our TSA.

We also conduct experiments by replacing our LWA with the concurrent DFM presented in DFMNet \cite{Zhang2021DFMNet}. We can observe that the performance significantly degrades. The difference between DFM and ours is in the manner to compute the similarity matrix. Specifically, DFM assumes a perfect alignment between multi-modalities and realizes a pixel-wise matrix multiplication, while we leverage the non-local attention with flattened vectors to compute the similarity. 

\begin{table}[t]
\footnotesize
\setlength\tabcolsep{0.5pt}
\setlength\extrarowheight{1pt}
\begin{center}
\caption{Ablation study on key components. $B$ stands for the baseline performance where RGB-D features are merged through simple addition without any form of attention.}
\label{tab:3dvablation}
\begin{tabular}{m{.64cm} m{.64cm} m{.64cm} m{.64cm} m{.64cm} m{.64cm}  m{.64cm} m{.64cm} m{.64cm} m{.64cm}  m{.64cm}  m{.64cm}}
\hline

\hline

 \multirow{2}{*}{$B$} & \multicolumn{4}{c}{CRM} & DFM  & LWA  & Size  & \multicolumn{4}{c}{Overall Metric}  \\

\cline{2-5} \cline{9-12}
& CA & TSA & SA  & $\alpha, \beta$ & (\cite{Zhang2021DFMNet}) &     & Mb$\downarrow$  & \small $M\downarrow$ & \small $F\uparrow$   & \small $S\uparrow$ & \small $E\uparrow $  \\
\hline
\checkmark & & & & & & & \textbf{305} &.039 &.915 &.904 &.935  \\
\checkmark & \checkmark & & & & & &336 & .035 &.918 &.907 &.940 \\
\checkmark & \checkmark & \checkmark & & & & & 364& .035 &.923 &.910 &.943 \\
\checkmark & \checkmark &  &\checkmark&  & & & 363& .035 &.920 &.908 & .941\\
\checkmark & \checkmark & \checkmark &  & \checkmark & & & 364 &.034 &\textbf{.924} &.910 &.943\\
\checkmark & \checkmark & \checkmark  & & \checkmark & \checkmark& &364 &.035 &.921 &.908 &.941\\ 
\checkmark & \checkmark & \checkmark &  &  \checkmark & &  \checkmark & 365 &\textbf{.033} &\textbf{.924} &\textbf{.911} &\textbf{.944}\\

\hline

\hline
\end{tabular}
\end{center}
\vspace{-6mm}
\end{table}

%
\section{Conclusion}

In this paper, we proposed a novel fusion architecture for RGB-D saliency detection. Different from previous works, we improve the robustness against inaccurate and misaligned depth inputs. Specifically, we proposed a novel layer-wise attention to explicitly leverage the depth quality by learning the best trade-off between early and late fusion. Furthermore, we improved the vanilla spatial attention to a broader context, yielding a simple yet efficient mechanism to address the depth misalignment problem. Extensive comparisons on benchmark datasets validate the effectiveness and robustness of our approach compared to the state-of-the-art alternatives. Our method also sets new records on challenging datasets with smaller model sizes. The method developed in this paper can potentially be used for other tasks, such as semantic segmentation and object detection, in a similar setting of RGB-D inputs in a robust manner.  

\section*{Acknowledgements}
This research is financed in part by the French Conseil R\'egional de Bourgogne-Franche-Comt\'e, the French National Research Agency under grant ANR CLARA (ANR-18-CE33-0004), and the French "Investissements d'Avenir" program ISITE-BFC (ANR-15-IDEX-0003).  
{\small
\bibliographystyle{ieee_fullname}
\bibliography{egbib}
}
\end{document}